\documentclass[conference]{IEEEtran}
\IEEEoverridecommandlockouts
\usepackage{cite}
\usepackage{amsmath,amssymb,amsfonts}
\usepackage{algorithmic}
\usepackage{graphicx}
\usepackage{textcomp}
\usepackage{xcolor}
\usepackage{multirow}
\def\BibTeX{{\rm B\kern-.05em{\sc i\kern-.025em b}\kern-.08em
    T\kern-.1667em\lower.7ex\hbox{E}\kern-.125emX}}
\begin{document}

\title{Multi-Task Text Classification using Graph Convolutional Networks for Large-Scale Low Resource Language}

\makeatletter
\newcommand{\linebreakand}{%
  \end{@IEEEauthorhalign}
  \hfill\mbox{}\par
  \mbox{}\hfill\begin{@IEEEauthorhalign}
}
\makeatother

\author{
\IEEEauthorblockN{Mounika Marreddy \textsuperscript{1} \IEEEauthorblockA{mounika.marreddy@research.iiit.ac.in}} \and
\IEEEauthorblockN{Subba Reddy Oota\textsuperscript{1} \IEEEauthorblockA{oota.subba@students.iiit.ac.in}} \and
\IEEEauthorblockN{Lakshmi Sireesha Vakada \textsuperscript{1} \IEEEauthorblockA{lakshmi.sireesha@research.iiit.ac.in}} \and 
\linebreakand
\IEEEauthorblockN{Venkata Charan Chinni\textsuperscript{1} \IEEEauthorblockA{venkata.charan@students.iiit.ac.in}} \and  
\IEEEauthorblockN{Radhika Mamidi\textsuperscript{1}
\IEEEauthorblockA{radhika.mamidi@iiit.ac.in}} \\ 
\linebreakand
\textsuperscript{1} IIIT Hyderabad 
}


\newcommand{\myspacing}{6pt}
\renewcommand\textfloatsep{\myspacing}
\renewcommand\floatsep{\myspacing}
\renewcommand\intextsep{\myspacing}
\renewcommand\dbltextfloatsep{\myspacing}
\renewcommand\dblfloatsep{\myspacing}

\maketitle

\begin{abstract}
Graph Convolutional Networks (GCN) have achieved state-of-art results on single text classification tasks like sentiment analysis, emotion detection, etc. However, the performance is achieved by testing and reporting on resource-rich languages like English. Applying GCN for multi-task text classification is an unexplored area. Moreover, training a GCN or adopting an English GCN for Indian languages is often limited by data availability, rich morphological variation, syntax, and semantic differences. In this paper, we study the use of GCN for the Telugu language in single and multi-task settings for four natural language processing (NLP) tasks, viz. sentiment analysis (SA), emotion identification (EI), hate-speech (HS), and sarcasm detection (SAR). 
In order to evaluate the performance of GCN with one of the Indian languages, Telugu, we analyze the GCN based models with extensive experiments on four downstream tasks.
In addition, we created an annotated Telugu dataset, \textsc{TEL-NLP}, for the four NLP tasks.
Further, we propose a supervised graph reconstruction method, Multi-Task Text GCN (MT-Text GCN) on the Telugu that leverages to simultaneously (i) learn the low-dimensional word and sentence graph embeddings from word-sentence graph reconstruction using graph autoencoder (GAE) and (ii) perform multi-task text classification using these latent sentence graph embeddings.
We argue that our proposed MT-Text GCN achieves significant improvements on \textsc{TEL-NLP} over existing Telugu pretrained word embeddings~\cite{marreddy2021clickbait}, and multilingual pretrained Transformer models: mBERT~\cite{devlin2018bertmulti}, and XLM-R~\cite{conneau2020unsupervised}. 
On \textsc{TEL-NLP}, we achieve a high F1-score for four NLP tasks: SA (0.84), EI (0.55), HS (0.83) and SAR (0.66).
Finally, we show our model's quantitative and qualitative analysis on the four NLP tasks in Telugu. 
We open-source our TEL-NLP dataset, pretrained models, and code~\footnote{https://github.com/scsmuhio/MTGCN\_Resources}.
\end{abstract}

\begin{IEEEkeywords}
Graph Convolutional Networks, Text Classification, Low resource language, Multi-Task, Indian Languages
\end{IEEEkeywords}

\section{Introduction}

\label{sec:intro}
Text classification is a well-studied problem in the field of NLP, which has important applications like sentiment analysis (SA)~\cite{liu2012sentiment,chen2018lifelong}, emotion identification (EI)~\cite{abdul2017emonet, tokuhisa2008emotion}, hate-speech detection (HS)~\cite{kshirsagar2018predictive,zhang2018detecting}, and sarcasm detection (SAR)~\cite{joshi2016word,joshi2017automatic}. Understanding the contextual and semantic representation of text is one of the main challenges in text classification. 
Also, the traditional methods represent text with hand-crafted features, including structural and part-of-speech (POS) tagging features, and depend on the availability of rich resources.

With the recent advancements in deep learning and the availability of text corpora for the English, neural networks like Convolutional Neural Networks (CNN)~\cite{kim2014convolutional} and Long-Short Term Memory Networks (LSTM)~\cite{hochreiter1997long} has been used for efficient text representation.
These models can capture semantic, syntactic information and provide a fixed-length vector (latent representation) by taking the sequence of words as input.
Additionally, these models have shown great success in the multi-task setting, where shared knowledge helps the model learn more general representation from the input text.
However, the input consecutive word sequences are local to these deep models, but they may ignore global word occurrence.  

Recently, a new research direction viz. Graph Neural Networks (GNN)~\cite{scarselli2008graph} or graph embeddings have attracted wider attention in many NLP tasks such as text classification~\cite{defferrard2016convolutional}, neural machine translation~\cite{bastings2017graph}, and relational reasoning~\cite{battaglia2016interaction}. 
GNN-based models adopt building a single large graph on the corpus that manages to learn reduced dimension representations and preserve the graph's global structural information, even with limited labeled data. 
With the recently proposed method, Graph Convolutional Networks (GCN) is used to generalize (convolutional) neural network models to perform text classification in English for a single task~\cite{yao2019graph}. 
However, applying a GCN for single or multi-task text classification for resource-poor Indian languages is unexplored. Moreover, adopting a GCN for Indian languages is often limited by data availability.
Hence, creating annotated data for multiple NLP tasks help the researchers to create better models for Telugu.


%
Unlike English, very few preparatory works have studied text classification in Telugu. Unfortunately, all these works were limited to existing baseline models that use either small annotated datasets~\cite{mukku2017actsa} or restricted to single task text classification~\cite{mukku2017tag}. Also, none of the works focused on learning the low-dimensional word or sentence graph representations from GCNs, which extracts the direct and indirect interactions between words and sentences in the graph embeddings. GCN-based approaches are alluring in this scenario as they perform both graph reconstruction (GAE) and node classification.

Inspired by the Graph AutoEncoder (GAE) model~\cite{schlichtkrull2018modeling} and Multi-Task Text classification~\cite{liu2016recurrent}, we propose a supervised graph reconstruction method, MT-Text GCN, an end-to-end model that jointly optimizes the reconstruction and multi-task text classification loss.
Further, our proposed method simultaneously learns word \& sentence representations for word-sentence graph reconstruction and sentence representations for multi-task text classification.
In summary, the main contributions of this paper are as follows:
\begin{itemize}
    \item We propose the multi-task learning model (MT-Text GCN) to reconstruct word-sentence graphs while achieving multi-task text classification with learned graph embeddings.
    \item \textsc{TEL-NLP} is the largest NLP dataset in Telugu covering four NLP tasks (16,234 samples for SA, HS, SAR, and 9,675 samples for EI),  unlike previous works, which focused only on a single task (SA) with a limited dataset of 2500 samples~\cite{mukku2017actsa}.
    \item For the Telugu language, we are the first to generate word embeddings using random walk based models: DeepWalk and Node2Vec, and Graph AutoEncoders (GAE).
    \item Our MT-Text GCN enabled better performance on Telugu when compared with multilingual pretrained transformer models on four NLP tasks.
\end{itemize}
    

\section{Related Work}
\noindent\textbf{GCN for Text Classification:}
In the current research, graph neural networks has attracted wide attention~\cite{battaglia2018relational}, and applied on various NLP tasks such as text classification~\cite{tayal2019short,yao2019graph}, semantic role labeling~\cite{marcheggiani2017encoding}, and machine translation~\cite{bastings2017graph}.
The earlier works in text classification using GCN focused with nodes being either words or documents to construct a single large graph~\cite{defferrard2016convolutional}.
Some authors construct a single graph by jointly embedding the documents and words as nodes (Text GCN)~\cite{yao2019graph} while considering the global relations between documents and words.
The main drawback of the existing GCN works is that they cannot capture the contextual word relationships within the document.
~\cite{zhang2020every} developed a model, TextING, for inductive text classification to overcome the above limitations.
~\cite{kipf2016semi} introduced GCN to embed the graph structure in a lower embedding space using neural networks.
Motivated by~\cite{kingma2013auto},~\cite{kipf2016variational} propose a graph AutoEncoder framework that uses GCN as an encoder to get the latent representation of each node and simple inner product as a decoder to reconstruct the graph.

\noindent\textbf{Deep Learning for Multi-Task Text Classification:}
There is a vast literature tackling the single-task text (ST-Text) classification using deep learning models such as CNNs~\cite{kim2014convolutional,zhang2015character} or RNNs~\cite{liu2016recurrent}.
However, these ST-Text classification approaches increase costs to build the resources and fail to combine multiple tasks. 
Multi-task learning can improve the classification performance of related tasks by learning these tasks in parallel~\cite{caruana1997multitask}. 
The literature survey shows that multi-task learning has been proven effective in solving similar NLP text classification problems~\cite{liu2016recurrent, xiao2018gated}.

\noindent\textbf{Text Classification for Telugu:}
Although extensive research happens in NLP tasks for the English, very few authors in literature worked on text classification (SA tasks) for Telugu language~\cite{mukku2017tag, parupalli2018enrichment, gangula2018resource}.
In~\cite{mukku2017tag, gangula2018resource}, authors provided Telugu SA data for different domains, categorized into three classes as ``positive'', ``negative'', and ``neutral''. 
Several authors use deep learning-based models such as CNN and LSTM to perform SA tasks on Telugu data in~\cite{varshit2018sentiment,tummalapalli2018towards}.
The authors in~\cite{mukku2017tag} used a semantic representation of each sentence using Telugu Word2Vec embeddings created on Wikipedia.
However, the corpus size is minimal (e.g., Telugu SA  dataset~\cite{mukku2017tag} has only 2500 samples), leading to relatively brittle systems.
Also, the authors use a 2X (two annotators perform the annotation) annotated dataset.
Recently,~\cite{choudhary2018sentiment} developed a deep learning-based model for Telugu SA to learn representations of resource-poor languages by jointly training with English corpora using a Siamese network. Here, the model trained on English - Hindi corpus and transfer learning is applied to the Telugu corpus, resulting in low accuracy.

\section{TEL-NLP Details}
\label{sec:dataset}
This section presents our dataset creation, \textsc{TEL-NLP}, for four NLP tasks such as SA, EI, HS, and SAR in Telugu. Here, we describe the dataset creation and annotation, inter-annotator agreement, and report the dataset statistics.

\noindent\textbf{Dataset Creation and Annotation} 
We sampled the 20k sentences from the existing large Telugu raw corpus~\cite{marreddy2021clickbait} and given to XYZ company to perform the annotation task.  
Also, the annotation was carried out for four NLP tasks viz. sentiment, emotion, hate-speech, and sarcasm.
We create a web-based user interface to simplify the annotation task.
Each user has to log in and do the annotation. 
The user interface shows the sentence and the respective labels for the four tasks. 
Five native speakers of Telugu with excellent fluency performed this annotation task. We provided the annotators with detailed definitions for each task with example sentences. All the annotators should read the sentence carefully and select the fitting label for that particular sentence. When the users click the submit button, their responses are saved in the database.
To qualify and complete the annotation process, each annotator must correctly annotate at least 80\% sentences from the sample dataset.
After two rounds of sanity checks, we obtained a total of 16,234 labeled sentences for SA, HS, SAR, and 9,675 for EI. Here we considered ``Positive'' and ``Negative'' labels for the SA task. ``Fear'', ``Angry'', ``Sad'', and ``Happy'' labels for EI task. ``Yes" and ``No" for HS and SAR datasets. As we considered the four basic emotions for the EI task, we got fewer samples when compared with other tasks.

\noindent\textbf{Inter-Annotator Agreement}
To perform the annotation, we made the five annotators work on a smaller dataset for verification as a first step. 
The Fleiss’ kappa score\footnote{https://en.wikipedia.org/wiki/Fleiss$\%$27$\_$kappa} was 0.95. 
The inter-annotator agreement on the whole dataset annotation was reported as 0.91

\noindent\textbf{TEL-NLP Description}
We performed our experiments on SA, EI, HS, and SAR datasets. The TEL-NLP dataset description and number of nodes in the graph are provided in Table~\ref{tab:dataset_stats}.

\begin{table}[t]
\scriptsize
\centering
\caption{Statistics of the \textsc{TEL-NLP} dataset.}
\label{tab:dataset_stats}
\resizebox{0.5\textwidth}{!}{\begin{tabular}{|l|c |c |c |c |}
\hline
\multirow{1}{*}{\textbf{Task}} & \multicolumn{1}{c|}{\# \textbf{Words}} & \multicolumn{1}{c|}{\# \textbf{Sentences}} & \multicolumn{1}{c|}{\# \textbf{Nodes}} & \multicolumn{1}{c|}{\# \textbf{Classes}}   \\ \cline{1-5} 
SA & 41501 & 16234 & 57735  & 2 (Positive: 9,488, Negative: 6,746)   \\ \hline
 &  &  &  & 4 (Fear: 121, Angry: 388) \\
EI &28447  & 9675 & 57735 &  (Sad: 5,135, Happy: 4,031) \\ \hline
HS & 41501 & 16234 & 57735 & 2 (Yes: 370, No: 15,864) \\\hline
SAR & 41501 & 16234 & 57735  & 2 (Yes: 402, No: 15, 832)  \\
\hline 
\end{tabular}}

\end{table}



\section{Proposed Method}
\label{sec:approach}
In this section, we will explain (i) the overview of Graph Convolutional Network (GCN)~\cite{kipf2016semi} and Graph AutoEncoder (GAE), (ii) the proposed MT-Text GCN approach, and (iii) the procedure to build word-word, sentence-sentence, and word-word + word-sentence graphs.

\subsection{GCN \& Graph AutoEncoder}
GCN is a multi-layer neural network that convolves a neighboring node's features and propagates its embedding vectors to its nearest neighborhoods.
Consider a weighted, undirected graph denoted by $\mathcal{G}$ : = ($\mathcal{V}$, A, X), where $\mathcal{V}$ is a set of N nodes, and A $\in \mathbb{R}^{NXN}$ is the symmetric adjacency matrix and X $\in \mathbb{R}^{NXM}$ is the node feature matrix with each node dimension M.
The Graph AutoEncoder~\cite{schlichtkrull2018modeling} (GAE) obtains a reduced dimensional representation (Z $\in \mathbb{R}^{NXK}$) of the input graph (A).
Intuitively, starting from the node embedding, the GAE model can reconstruct an adjacency matrix $A^{'}$ close to the empirical graph (A), then the Z vectors should capture some essential characteristics of the graph structure.
To reconstruct the graph, we stack an inner product decoder to this GAE as follows:
\setlength{\belowdisplayskip}{0pt} \setlength{\belowdisplayshortskip}{0pt}
\setlength{\abovedisplayskip}{0pt} \setlength{\abovedisplayshortskip}{0pt}
\begin{align}
   \label{gae}
   A^{'} = f(AH^{(1)}W_{1}) \\  
    Z = H^{(1)} = f(AXW_{0})
\end{align}
where \emph{f} denotes the activation function, W$_{0}$ and W$_{1}$ denotes the weight parameters learned from the encoder and decoder model, respectively.

\subsection{Multi-Task Text GCN (MT-Text GCN)}
Given an input graph A, our main objective is to build and train a model to reconstruct a graph (GAE) to learn word and sentence embeddings. These learned embeddings are used to perform multi-task text classification.
Here, we perform multi-task text classification by providing sentence representations in two ways : (i) learned sentence embeddings from GAE, (ii) each sentence is represented by taking the average of learned word embeddings from GAE. The proposed MT-Text GCN pipeline is shown in
Fig~\ref{fig:MT-Text GCN}.
We use a 2-layer GCN of which layer-1 acts as encoder and layer-2 as the decoder.
The encoder (Z) and decoder output ($A^{'}$) using the layer-1 latent representations can be estimated from the Equations~\ref{gae} and ~\ref{pme}.

We use a LightGBM classifier to perform the multi-task text classification.
Overall, the final objective function is given as
\setlength{\belowdisplayskip}{0pt} \setlength{\belowdisplayshortskip}{0pt}
\setlength{\abovedisplayskip}{0pt} \setlength{\abovedisplayshortskip}{0pt}
\begin{equation}
    \mathcal{L} = \mathcal{L}_{MSE}(A, A^{'}) + \lambda \times \mathcal{L}_{MT_{CLA}}(\widehat{l}, l)
\end{equation}
\noindent where, $\mathcal{L}_{MSE}$ (reconstruction loss) denotes the mean squared error (MSE) between the reconstructed graph $A^{'}$ and the empirical graph $A$. The sigmoid cross-entropy loss denoted by $\mathcal{L}_{MT_{CLA}}$ evaluate classification performance between predicted labels $\widehat{l}$ and ground truth labels (l).
The $\lambda$ parameter tunes the trade-off between reconstruction and classification performance. The model performance for different values of $\lambda$ are reported in Fig~\ref{fig:gcn-loss}.  

\begin{figure}[t]
\centering
    \includegraphics[width=\linewidth]{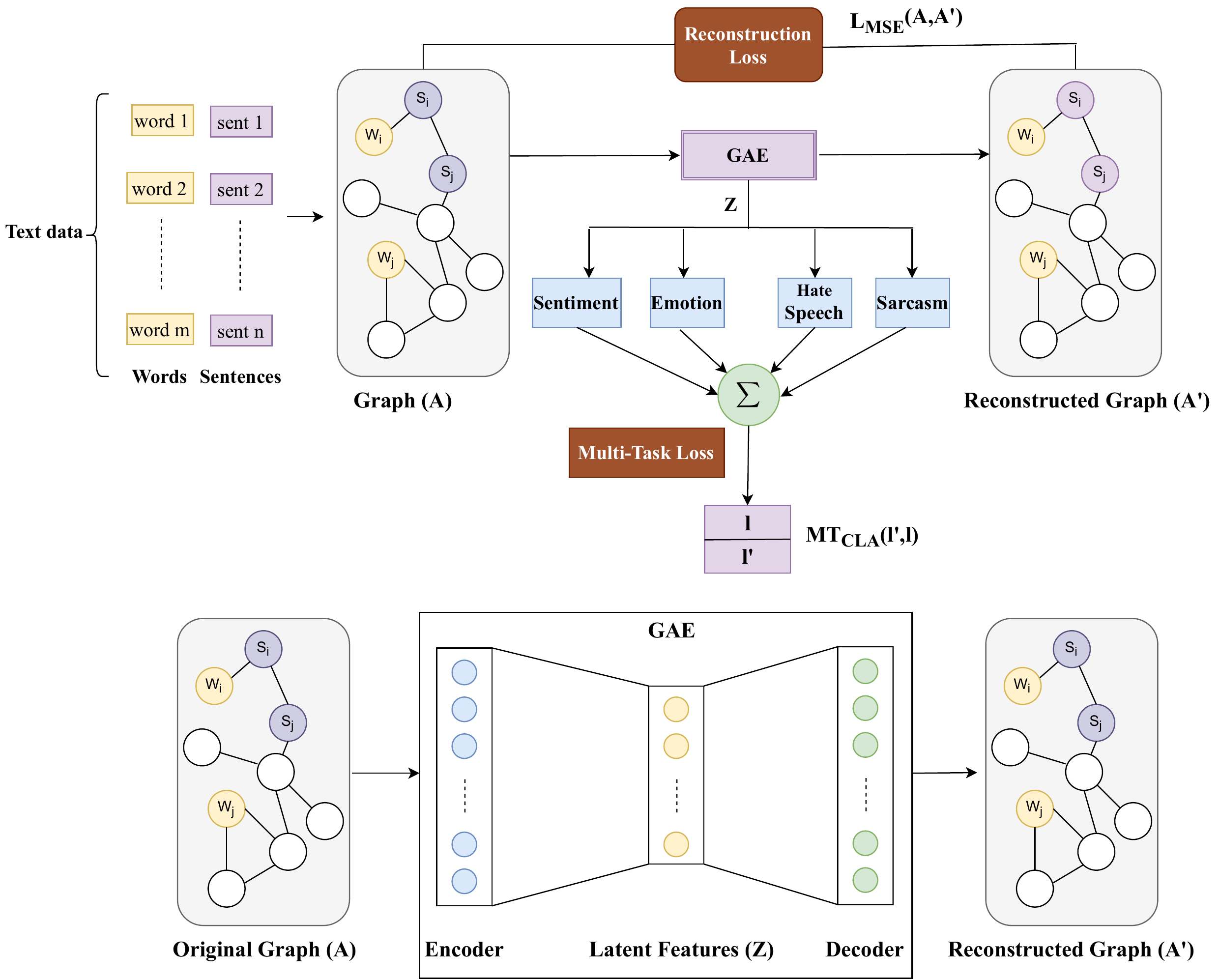}
    \caption{MT-Text GCN Pipeline: Rows in Z are low-dimensional node embeddings learned by the encoder from input graph A. LightGBM classifier outputs predicted labels $\widehat{l}$, and compares with actual labels l. $\mathcal{L}_{MSE}$ (reconstruction loss) denotes the mean squared error (MSE) between the reconstructed graph $A^{'}$ and the empirical graph $A$. The sigmoid cross-entropy loss denoted by $\mathcal{L}_{MT_{CLA}}$}
    \label{fig:MT-Text GCN}
\end{figure}


\subsection{Building Text Graph}
We set the node feature vector (X) as an identity matrix, i.e., every word/sentence/word+sentence represents a one-hot vector, given as input to MT-Text GCN.

\noindent\textbf{Word-level Graph (W):}
To construct a word-level graph, we explicitly calculated the global word co-occurrence matrix with a window size of 3, and the number of nodes is equal to the number of unique words in the vocabulary.
Since point-wise mutual information (PMI)~\cite{damani2013improving} yields better results than word co-occurrence count~\cite{yao2019graph}, we adopt PMI as a metric to calculate the weights between any two words.
We use the global word-word co-occurrence matrix as the adjacency matrix (A), where the edge between the two words is calculated using PMI.
\setlength{\belowdisplayskip}{0pt} \setlength{\belowdisplayshortskip}{0pt}
\setlength{\abovedisplayskip}{0pt} \setlength{\abovedisplayshortskip}{0pt}
\begin{equation}
\label{pme}
    PMI(i, j) = log\frac{p(i, j)}{p(i)p(j)}
\end{equation}
\begin{equation}
    p(i, j) = \frac{\#W(i, j)}{\#W}  
\end{equation}
\begin{align}
    p(i) &= \frac{\#W(i)}{\#W}
\end{align}
where \#W(i, j) is the number of sliding windows that contain words i and j, \#W(i) is the number of sliding windows that contain the word i, \#W is the total number of sliding windows in the dataset.

\noindent\textbf{Sentence-level Graph (S):}
In this graph, nodes represent the sentences, and we use pretrained word embeddings (Word2Vec) to obtain sentence vector representation by averaging all the word vectors of the sentence.
We use cosine distance to calculate the similarity score between two nodes. 

\noindent\textbf{Word+Sentence-level Graph (W+S):}
In this graph, words and sentences act as nodes, we adopt the global word co-occurrence statistics and TF-IDF frequencies to fill the edge weights between the nodes.
\begin{equation}
    A_{ij} = \begin{cases}
    \text{PMI}(i, j) ,& \text{i, j are words}\\
    \text{TF-IDF}_{ij} ,& \text{i is word, j is sentence} \\
    1,& i=j \\
    0,              & \text{otherwise}
\end{cases}
\end{equation}

\section{Experimental Setup \& Results}
\label{sec:exp}
This section describes different features such as traditional features, pretrained word embeddings, random walk-based representations, and MT-Text GCN model training setup. 
This paper uses Light Gradient Boosting Method (LightGBM) as a classifier to train the models for all the feature representations.

\subsection{Baselines}
We compare our MT-Text GCN with traditional features, distributed word representations, and random walk based methods.

\noindent\textbf{Traditional Features}
We use the traditional Bag-of-words (BoW)~\cite{wallach2006topic} and TF-IDF features~\cite{ramos2003using} as our baseline methods to perform text classification on four NLP tasks.
The BoW model uses the frequency of words in a sentence as features. The BoW model with term frequency times inverse document frequency of words in a document is used as TF-IDF features.

\subsection{Telugu Word Embeddings} 
We use the existing pretrained Telugu word embeddings~\cite{marreddy2021clickbait} to perform the text classification on our dataset,~\textsc{TEL-NLP}:
(1) A \textbf{Te-Word2Vec (Te-W2V)} model trained on a large Telugu corpus with 300-hidden dimensions as word representations. 
(2) \textbf{Te-GloVe} based word vectors (each word is a 200-dimension vector).
(3) A 200-dimension \textbf{Te-FastText (Te-FT)} embeddings.
(4) \textbf{Te-Meta-Embeddings (Te-ME)}, an ensemble of Telugu Te-W2V, Te-GloVe, and Te-FT embeddings are used to create TeMe of 300 dimension~\cite{marreddy2021clickbait}.

\subsection{Deep Learning Methods}
\noindent\textbf{CNN:} 
Proposed by~\cite{kim2014convolutional}, we perform convolution and max-pooling operations on Telugu word embeddings to get a representation of text that is used for text classification.

\noindent\textbf{LSTM:} Defined in~\cite{tang2016effective, liu2016recurrent}, we use the last hidden state as text representation to perform text classification. We train LSTM with five input representations (i) One-Hot encoding and (ii) the four pretrained Telugu word embeddings~\cite{marreddy2021clickbait}.

\subsection{Random Walk based Representations}
\noindent\textbf{DeepWalk (DW):} Proposed by~\cite{perozzi2014deepwalk}, this approach uses local information obtained from truncated continuous feature representations random walks, equivalent to sentences to learn latent representations of nodes in a graph. 

\noindent\textbf{Node2Vec (N2V):} A framework for learning nodes in the graph by using a biased random walk procedure, which efficiently explores diverse neighborhoods in~\cite{grover2016node2vec}. 
To accommodate the morphological richness in Indian languages like Telugu, we consider the different small window sizes (2,3,4) for creating the co-occurrence matrices.

\begin{figure}[t]
    \centering
    \small
    \setlength{\tabcolsep}{2pt}
    \begin{tabular}{|c|}
    \hline
    \includegraphics[width=0.7\linewidth]{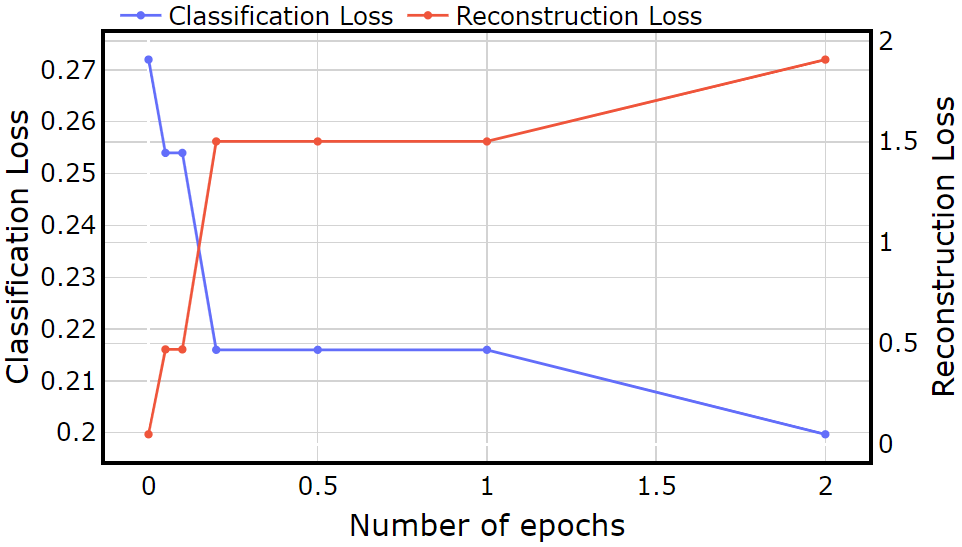} \\ 
         \includegraphics[width=0.7\linewidth]{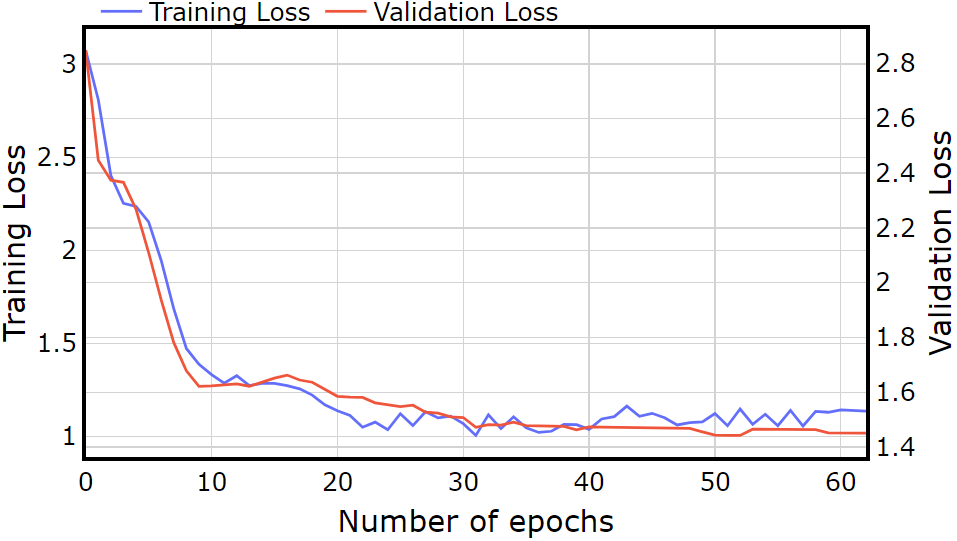} \\ \hline
    \end{tabular}
    \caption{Example of Reconstruction (red)/ classification (blue) loss with different $\lambda$ (top), Train and Validation Learning Curves Shows a Good Fit at $\lambda$ = 0.2 (bottom). Each point on the curve is the average of loss with corresponding $\lambda$.}
    \label{fig:gcn-loss}
\end{figure}

\subsection{Graph Based Representations}
\noindent\textbf{GCN-W:}
A word-level graph CNN model that operates convolution over word-word adjacency matrix~\cite{yao2019graph}, in which first-order approximation of Chebyshev filter is used.

\noindent\textbf{GCN-S:} 
A sentence-level graph CNN model that operates convolution over sentence embeddings similarity matrix~\cite{zhao2020summpip}, in which first-order approximation of Chebyshev filter is used.

\noindent\textbf{GCN-W+S:}
A word+sentence level graph CNN model that operates convolution over word-word and word-sentence embeddings similarity matrix~\cite{yao2019graph}, in which first-order approximation of Chebyshev filter is used.

\subsection{Model Setup}
To perform text classification using GCN, we set the first convolution layer's embedding size as 200 and the input feature vector as a one-hot vector for different GCN models such as GCN-W, GCN-S, GCN-W+S, and MT-Text GCN.
We use the Adam optimizer~\cite{kingma2014adam} with an initial learning rate of 0.001, RELU as the activation function at the first GCN layer, and L2 weight decay equal to 5e$^{-4}$.
We applied the dropout with a probability of 0.5 after the first GCN layer and trained all the GCN models for a maximum of 100 epochs.
We set the callbacks and checkpoints in which the model training stops automatically if the validation loss does not decrease for ten consecutive epochs.
For the GCN-W model, we set the context window size as three while constructing a word-word co-occurrence matrix using PMI as an input adjacency matrix.
Here, we jointly perform the reconstruction of a graph using GAE and the latent node representations from GAE to build the classification model.
The experiments were performed on a single V100 16GB RAM GPU machine.

To determine the choice of $\lambda$, we perform an extensive grid search
and the corresponding results are shown in Fig~\ref{fig:gcn-loss}.
As the $\lambda$ increases, the reconstruction performance deteriorates, and classification loss improves. 
When $\lambda$=0, the reconstruction loss yields the best score, whereas classification loss is peak.
In the end, we chose $\lambda$ = 0.2 based on the performance of both reconstruction and classification loss.
Fig~\ref{fig:gcn-loss} depicts the model training vs. validation loss performance when passing the parameter $\lambda$ =0.2. Here the first layer embedding dimensions are 200.
Fig~\ref{fig:gcn-loss} tells us that both training and validation loss decreased as the number of epochs increased.
We also experimented with GCN-W and GCN-S on a single task for different embedding dimensions, and corresponding F1-scores for each model are shown in Table~\ref{tab:CB_results}.

\noindent\textbf{Dataset Splitting:}
We followed the 5-fold cross-validation split for all the methods. One fold was used for the test (20\% data), and the 4-fold data was split into train/validation (70/10).
We calculated the average of 5-folds and reported the results in Table~\ref{tab:CB_results}.

\noindent\textbf{Classification Metrics:}
Table~\ref{tab:CB_results} reports each task classification results such as F1-score for various models, except for the HS task.
Since our HS task data has a class imbalance issue, we report the weighted F1-score.

\section{Results and Analysis}
The effectiveness of our proposed model is evaluated by comparing with traditional features, different word embeddings~\cite{marreddy2021clickbait}, deep learning methods, node embeddings, and existing pretrained multilingual transformer models mBERT~\cite{devlin2018bertmulti} and XLM-R~\cite{conneau2020unsupervised}. The results can be analyzed from the Table~\ref{tab:CB_results} as follows:

\noindent\textbf{Traditional Features:}
We considered the traditional feature representations BoW and TF-IDF generated from the Telugu corpus as the baselines.
Since our baseline features do not capture the semantic and contextual information, the baseline system achieves lower F1-scores for all tasks, as shown in Table~\ref{tab:CB_results}, block (A).


\noindent\textbf{Distributed Word-Embeddings:}
For each text classification task, we used the pretrained word-embeddings generated from Telugu corpus (Te-W2V, Te-GloVe, Te-FT, and Te-ME)~\cite{marreddy2021clickbait} as input, and results are reported in Table~\ref{tab:CB_results}.
With the local (Te-W2V, Te-FT) and global (Te-GloVe) contextual word embeddings, the system achieves an improved F1-score compared to the baseline model for all the four NLP tasks in which Te-FT embeddings has shown better performance than other word-embeddings, as shown in Table~\ref{tab:CB_results}, block (B).
However, from Table~\ref{tab:CB_results}, we observe that the performance of sarcasm task results is deficient and equal to random guessing due to the high class imbalance problem. 

\setlength{\tabcolsep}{4pt}
\begin{table}[t]
\scriptsize
\centering
\caption{\textsc{TEL-NLP} results on four tasks in terms of F1-score using: (A) traditional features, (B) distributed word embeddings, (C) deep learning representations, (D) random walk embeddings, (E) pretrained multi-lingual transformers, and (F) our proposed method: MT-Text GCN and its variations. All these methods were trained on Telugu corpus. Since HS data has a class imbalance issue, we report the weighted F1-score.}
\label{tab:CB_results}
\begin{tabular}{|c|l|c| c|c|c| }
\hline
 &\multirow{1}{*}{Tasks$\rightarrow$} & \multicolumn{1}{c|}{SA} & \multicolumn{1}{c|}{EI} & \multicolumn{1}{c|}{HS} & \multicolumn{1}{c|}{SAR}   \\ \hline
Block & Feature set$\downarrow$ & F1-score  & F1-score  & F1-score &   F1-score\\ \hline 
(A) & BoW  & 0.60   & 0.32  & 0.56  & 0.50  \\ 
 & TF-IDF  & 0.61  & 0.32  & 0.58&  0.50 \\ \hline
& Te-W2V&   0.79   & 0.44   & 0.63  & 0.50    \\ 
(B) & Te-GloVe&   0.79  & 0.43  & 0.68  & 0.52 \\
& Te-FT&  0.83  & 0.50  & 0.66  & 0.55\\
& Te-ME & 0.82  & 0.48  & 0.68  & 0.50  \\ \hline
& CNN & 0.64   & 0.30   & 0.63   & 0.51 \\ 
& LSTM& 0.81 & 0.45   & 0.71  & 0.53   \\ 
& LSTM-TeW2V & 0.75 & 0.49   & 0.68   & 0.52   \\
(C) & LSTM-TeGloVe & 0.66  & 0.35   & 0.67   & 0.51   \\ 
& LSTM-TeFT & 0.72  & 0.41  & 0.67  & 0.53   \\ 
& LSTM-TeME & 0.65  & 0.38  & 0.68   & 0.52   \\ \hline

& DW & 0.56  & 0.30   & 0.70  & 0.50 \\ 
& N2V-WS2 & 0.77  & 0.48 & 0.66  & 0.54 \\ 
(D) & N2V-WS3 & 0.78 & 0.47  & 0.80  & 0.55 \\ 
& N2V-WS4 & 0.78 & 0.46 & 0.82  & 0.58 \\ \hline
(E) & mBERT & 0.68   & 0.42   &0.50  & 0.50    \\ 
& XLM-R & 0.64  & 0.34  &0.50   & 0.50   \\ 
\hline
& ST-TextGCN-W  & 0.69  & 0.36   & 0.75   & 0.64 \\ 
(F) & ST-Text GCN-S & 0.77  & 0.43   & 0.82  & 0.65 \\ 
& ST-Text GCN-W+S & 0.82   & 0.48  & 0.83  & \textbf{0.67} \\ 
& MT-Text GCN-W+S &   \textbf{0.84}   & \textbf{0.55}  & \textbf{0.83}   & 0.63 \\
\hline

\end{tabular}

\end{table}

\noindent\textbf{Deep Learning Methods:}
Although the deep learning models such as CNN and LSTM generated from Telugu corpus outperform the baseline setting, CNN with randomly initialized word embedding yields a lower F1-score when compared with different word embeddings, displayed in Table~\ref{tab:CB_results}, block(C). 
The LSTM-based model outperforms the traditional features and performs better than CNN.
Moreover, the sequence of tokens with one-hot representation as input to LSTM yields a better F1-score than four pretrained word-embeddings, except for the EI task.
These results infer that LSTMs do not address the long-term dependency problems in morphologically rich languages, Telugu, which affects the classification performance of four tasks.

\begin{figure*}[t]
\centering
    \includegraphics[width=\linewidth]{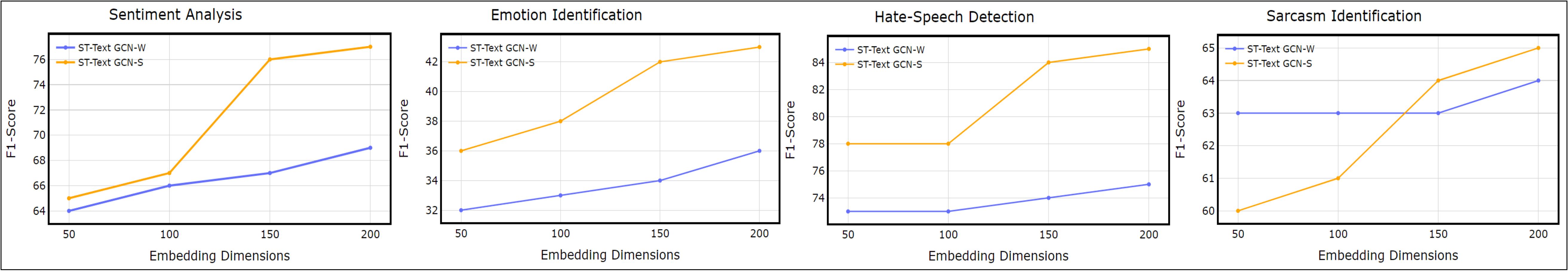}
    \caption{Test accuracy results for four tasks (i) SA, (ii) EI, (iii) HS, and (iv) SAR by varying embedding dimensions.}
    \label{fig:gcn-sen-emo}
\end{figure*}

\begin{figure}[t]
\centering
    \includegraphics[width=\linewidth]{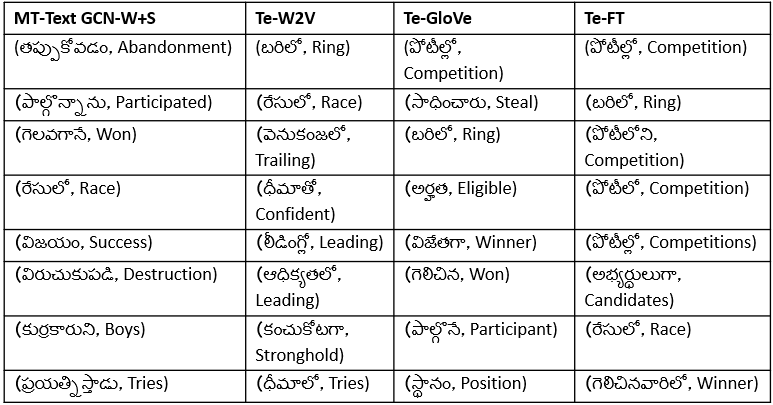}
    \caption{ Top 8 similar words for the word ``competition'' using (i) MT-Text GCN-W+S, (ii) Te-W2V, (iii) Te-GloVe, and (iv) Te-FT.}
    \label{fig:tsne}
\end{figure}

\noindent\textbf{Random Walk based Approaches:}
From Table~\ref{tab:CB_results}, block(D) describes the results of random walk based models (DW / N2V) generated from Telugu corpus as input to the LightGBM model for four tasks.
We make the following observations from Table~\ref{tab:CB_results}: (i) the DW embeddings reports a low F1-score for four tasks when compared to all the methods except baseline setting in Table~\ref{tab:CB_results}.
(ii) We observe that the N2V embeddings learned from the three context graphs (i.e., varying the different context window sizes: 2, 3, and 4) given as an input, the model displays an increasing F1-score and reports better results with a context window size of 3 \& 4.
(iii) Compared to distributed word embeddings and deep learning methods, N2V-WS3 and N2V-WS4 display higher results in F1 scores for the HS and SAR tasks.
(iv) These results demonstrate that N2V is ideal for learning both local and global neighbors so that it can construct more semantic vertex neighborhoods, and it is better at classifying the text classification tasks on the imbalanced datasets (HS and SAR), while DW has the limitation of no control over explored neighborhoods.

\noindent\textbf{Pretrained Transformers:}
We evaluate whether fine-tuning the existing pretrained multilingual transformer models mBERT~\cite{devlin2018bertmulti}, and XLM-R~\cite{conneau2020unsupervised} is useful for adapting each of the NLP tasks.
From Table~\ref{tab:CB_results}, we find that fine-tuning results show better performance than baseline, CNNs, and DW methods.
However, the performance is lower compared to Telugu pretrained word embeddings, N2V embeddings, and our proposed MT-Text GCN reported in Table~\ref{tab:CB_results}, block(E). 
This is because the average length of sentences in these Telugu NLP tasks is much longer than that in English datasets. Moreover, the graph is constructed using word-sentence statistics, meaning long texts may produce more sentence connections transited via an intermediate word node. This potentially benefits graph mechanisms, leading to better performances with MT-Text GCN multi-lingual Transformers.

\begin{figure*}[t]
\centering
    \includegraphics[width=0.88\linewidth]{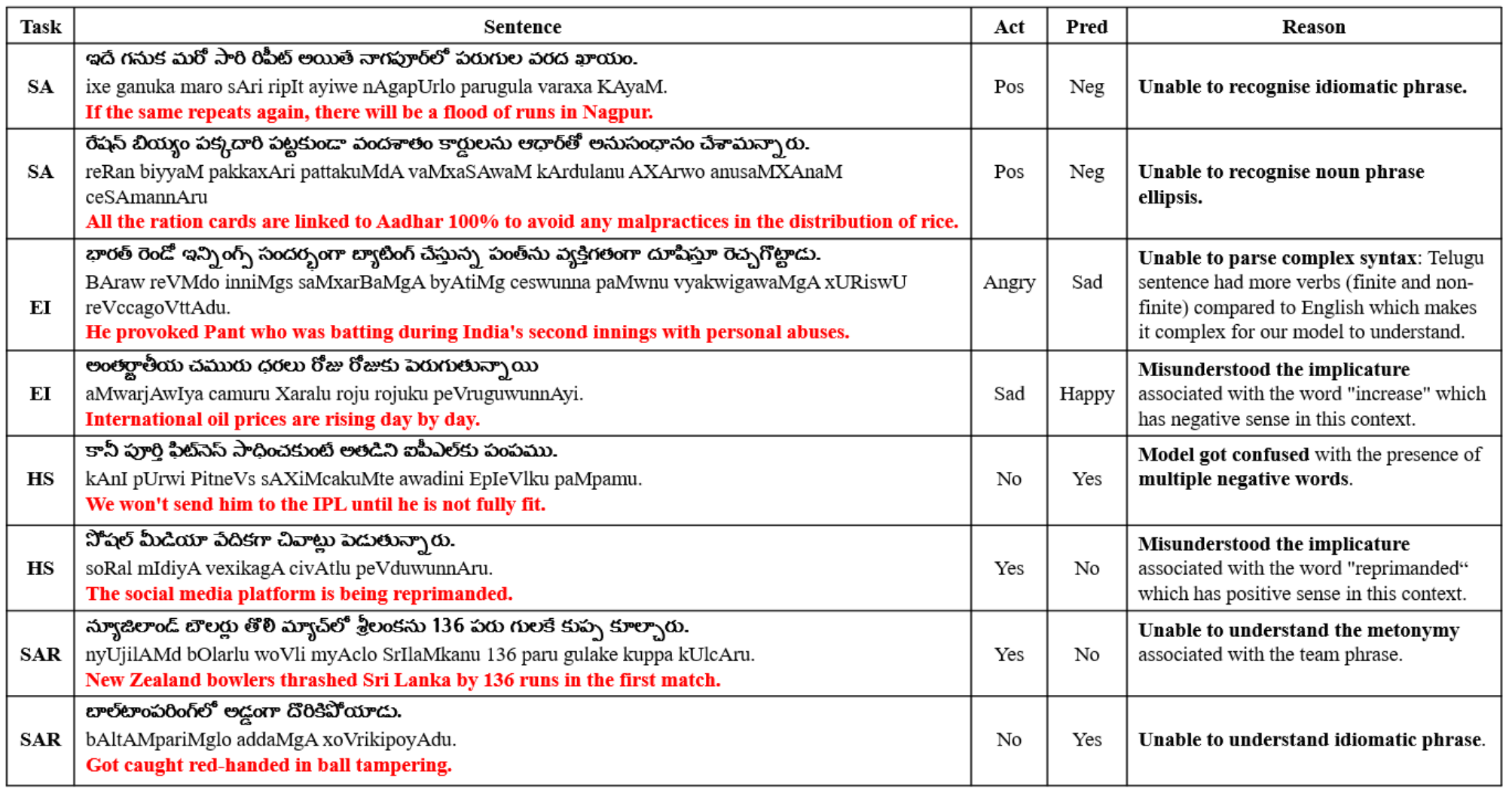}
    \caption{MT-Text GCN-W+S: Wrong predictions by the model on four text classification tasks: SA, EI, HS, SAR.}
    \label{fig:mtgcn_erros}
\end{figure*}

\subsection{Graph based Node Embeddings:}

\noindent\textbf{ST-Text GCN Results:}
We present the single task text classification results for models trained on word-word (ST-Text GCN-W), sentence-sentence (ST-Text GCN-S), and word-word + word-sentence (ST-Text GCN-W+S) graphs in Table~\ref{tab:CB_results}, block(F).
(i) From Table~\ref{tab:CB_results}, we infer that sentence embeddings learned from ST-Text GCN-S show a better classification performance across four individual tasks compared to word embeddings obtained from ST-Text GCN-W.
(ii) Further, the single task text classification model trained on the W+S graph achieved high performance compared to ST-Text GCN-W and ST-Text GCN-S.
(iii) These results indicate that graph-based models learned better contextual representation for each sentence rather than sentence representation obtained from an average of word embeddings.
(iv) Unlike ST-Text GCN-S, we are not providing any edge information between a sentence to sentence. However, we observe that the latent sentence embeddings learned from ST-Text GCN-W+S reports better than ST-Text GCN-S in Table~\ref{tab:CB_results}.
(v) Although EI and SAR tasks have a data imbalance problem, all the GCN models yield better results than other feature representations.
We reported that the quantitative analysis of each task is displayed in Fig ~\ref{fig:gcn-sen-emo}.

\noindent\textbf{MT-Text GCN Results:}
Like ST-Text GCN-W+S, we use the sentence embeddings as input to the classification model when training the MT-Text GCN-W+S.
Table~\ref{tab:CB_results} shows that (i) the classification performance of SA and HS tasks have increased the performance in the multi-task setting.
(ii) The classification performance of the SAR task was deficient, possibly due to class imbalance and lack of solid correlations across some task pairs.
(iii) Overall, from Table~\ref{tab:CB_results}, we observe that GCN-based models outperform all the tasks except a little less for the SAR task.

\subsection{Qualitative Analysis}
In this study, we tried to understand how MT-Text GCN helps in regularizing word embeddings.
Since word vectors are numerical representations of contextual similarities between the words, they can be manipulated and made to perform unusual tasks like finding the degree of similarity between two words~\cite{mikolov2013distributed}.
To verify the quality of generated Telugu word-embeddings, we extract the top-8 semantically related words for “(Competition)” using Te-W2V, Te-GloVe, Te-FT, and compared with MT-Text GCN-W+S, as shown in Fig~\ref{fig:tsne}.
From Fig~\ref{fig:tsne}, we can observe that the semantically related words obtained for MT-Text GCN-W+S are similar to Te-W2V, Te-GloVe, and Te-FT.

\subsection{Quantitative Analysis}
Fig~\ref{fig:gcn-sen-emo} showcase the classification performance (F1-score) of the models ST-Text GCN-W and ST-Text GCN-S on SA, EI, HS, and SAR datasets with different embedding dimensions at the first GCN layer. 
From Fig~\ref{fig:gcn-sen-emo}, we can observe that ST-Text GCN-S shows better classification performance and an increasing F1-score with different embedding dimensions when compared to ST-Text GCN-W.
These results infer that the increase in the dimension yields an improved accuracy over all the tasks.

\begin{figure}[t]
\centering
    \includegraphics[width=0.9\linewidth]{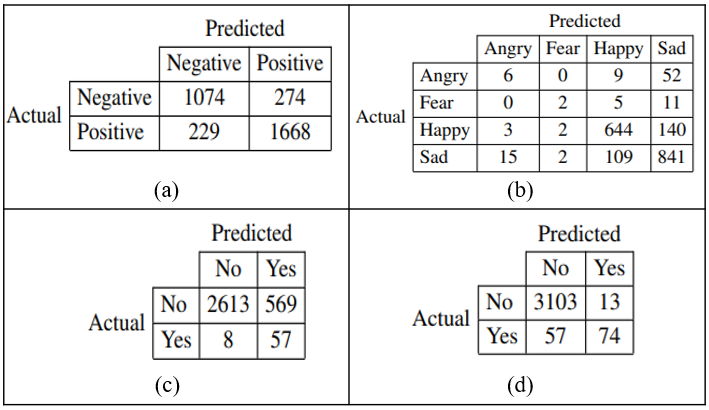}
    \caption{Confusion Matrices of MT-Text GCN-W+S for three best performing tasks: (a) SA, (b) EI, (c) HS, and (d) SAR using ST-Text GCN-W+S.}
    \label{fig:confusionMatrix-elmo}
\end{figure}

\subsection{Error Analysis}
We analyzed the error cases in detail for the class labels that belong to each task.
We observed from Fig~\ref{fig:confusionMatrix-elmo}a that for SA task: (i) 20\% of the Negative class samples are predicted as Positive,  and (ii) 12\% of Positive class samples are predicted as Negative.
For EI, we observe from the Fig~\ref{fig:confusionMatrix-elmo}b that (i) 78\% of Angry class samples, 65\% of Fear class samples, 17\% of Happy class samples are predicted Sad, and (ii) 11\% of Sad class samples are predicted Happy.
These results infer that our system is more biased towards the Sad class samples due to the class imbalance problem in Angry and Fear class samples.
Fig~\ref{fig:confusionMatrix-elmo}c shows the confusion matrix for the HS task, and we make the following observations: (i) The recall of Yes (88\%) and No (82\%) classes are much higher, (ii) our model makes a higher percentage of false negatives (i.e., 18\% of No class samples are predicted Yes). This is mainly due to the class imbalance issue, and we report the weighted F1-score for HS task data in results Table~\ref{tab:CB_results}. 
For the SAR task, we observe that (i) 43\% of Yes class samples are predicted No (i.e., our system makes more false positives), and (ii) the presence of false negatives (0.4\%) is low compared to the HS task, as shown in Fig~\ref{fig:confusionMatrix-elmo}d. 

Fig~\ref{fig:mtgcn_erros} displays the failure cases for four tasks, and we make the following observations:
(i) The model could not recognize ``idiomatic'' and ``noun phrase ellipses'' in the SA task.
(ii) For the EI task, the model misunderstood the word ``increase'' in a positive sense in this context and was unable to parse sentences with ``complex syntax''. For instance, sentence 3 in Fig~\ref{fig:mtgcn_erros} shows that the relative clause of English gets translated to an adjectival construction, and the noun phrase in the prepositional phrase gets verbalized in Telugu. Moreover, the Telugu sentence had more verbs (finite and non-finite) than English, making it complex for our model to understand.
(iii) In the HS task, the model got confused with positive words and also misunderstood the word ``reprimanded'' in a negative sense in this context.
(iv) Like the SA task, the model could not recognize ``idiomatic'' or ``metonymy'' in the SAR task.

\section{Conclusion}
In this paper, we propose an MT-Text GCN to jointly (i) perform graph reconstruction, (ii) learn word embeddings, (iii) learn sentence embeddings, and (iv) do multi-task text classification on four significant NLP tasks - SA, EI, HS, and SAR. The experimental results show that our proposed model reports an improved classification performance across three tasks. 
In the future, we aim to create pretrained transformer models for Telugu. 
We believe that this work will pave the way for better research in resource-poor languages. 

\section{Ethical Statement}
We reused publicly available information from ~\cite{marreddy2021clickbait} to create datasets for this work. We have gone through the privacy policy of various websites mentioned in the paper. For example, the greatandhra website privacy policy is provided here \footnote{https://www.greatandhra.com/privacy.php}. We do not foresee any harmful uses of using the data from these websites.

\bibliography{anthology,custom}

\begin{thebibliography}{10}
\providecommand{\url}[1]{#1}
\csname url@samestyle\endcsname
\providecommand{\newblock}{\relax}
\providecommand{\bibinfo}[2]{#2}
\providecommand{\BIBentrySTDinterwordspacing}{\spaceskip=0pt\relax}
\providecommand{\BIBentryALTinterwordstretchfactor}{4}
\providecommand{\BIBentryALTinterwordspacing}{\spaceskip=\fontdimen2\font plus
\BIBentryALTinterwordstretchfactor\fontdimen3\font minus
  \fontdimen4\font\relax}
\providecommand{\BIBforeignlanguage}[2]{{%
\expandafter\ifx\csname l@#1\endcsname\relax
\typeout{** WARNING: IEEEtran.bst: No hyphenation pattern has been}%
\typeout{** loaded for the language `#1'. Using the pattern for}%
\typeout{** the default language instead.}%
\else
\language=\csname l@#1\endcsname
\fi
#2}}
\providecommand{\BIBdecl}{\relax}
\BIBdecl

\bibitem{marreddy2021clickbait}
M.~Marreddy, S.~R. Oota, L.~S. Vakada, V.~C. Chinni, and R.~Mamidi, ``Clickbait
  detection in telugu: Overcoming nlp challenges in resource-poor languages
  using benchmarked techniques,'' in \emph{2021 International Joint Conference
  on Neural Networks (IJCNN)}.\hskip 1em plus 0.5em minus 0.4em\relax IEEE,
  2021, pp. 1--8.

\bibitem{devlin2018bertmulti}
J.~Devlin, M.-W. Chang, K.~Lee, and K.~Toutanova, ``Multilingual bert -r,''
  \emph{https://github.com/google-research/bert/blob/master/multilingual.md},
  2018.

\bibitem{conneau2020unsupervised}
A.~Conneau, K.~Khandelwal, N.~Goyal, V.~Chaudhary, G.~Wenzek, F.~Guzm{\'a}n,
  {\'E}.~Grave, M.~Ott, L.~Zettlemoyer, and V.~Stoyanov, ``Unsupervised
  cross-lingual representation learning at scale,'' in \emph{Proceedings of the
  58th Annual Meeting of the Association for Computational Linguistics}, 2020,
  pp. 8440--8451.

\bibitem{liu2012sentiment}
B.~Liu, ``Sentiment analysis and opinion mining,'' \emph{Synthesis Lectures on
  Human Language technologies}, vol.~5, no.~1, pp. 1--167, 2012.

\bibitem{chen2018lifelong}
Z.~Chen, N.~Ma, and B.~Liu, ``Lifelong learning for sentiment classification,''
  \emph{arXiv preprint arXiv:1801.02808}, 2018.

\bibitem{abdul2017emonet}
M.~Abdul-Mageed and L.~Ungar, ``Emonet: Fine-grained emotion detection with
  gated recurrent neural networks,'' in \emph{Proceedings of the 55th Annual
  Meeting of the Association for Computational Linguistics (volume 1: Long
  papers)}, 2017, pp. 718--728.

\bibitem{tokuhisa2008emotion}
R.~Tokuhisa, K.~Inui, and Y.~Matsumoto, ``Emotion classification using massive
  examples extracted from the web,'' in \emph{Proceedings of the 22nd
  International Conference on Computational Linguistics-Volume 1}.\hskip 1em
  plus 0.5em minus 0.4em\relax Association for Computational Linguistics, 2008,
  pp. 881--888.

\bibitem{kshirsagar2018predictive}
R.~Kshirsagar, T.~Cukuvac, K.~McKeown, and S.~McGregor, ``Predictive embeddings
  for hate speech detection on twitter,'' \emph{arXiv preprint
  arXiv:1809.10644}, 2018.

\bibitem{zhang2018detecting}
Z.~Zhang, D.~Robinson, and J.~Tepper, ``Detecting hate speech on twitter using
  a convolution-gru based deep neural network,'' in \emph{European Semantic Web
  Conference}.\hskip 1em plus 0.5em minus 0.4em\relax Springer, 2018, pp.
  745--760.

\bibitem{joshi2016word}
A.~Joshi, V.~Tripathi, K.~Patel, P.~Bhattacharyya, and M.~Carman, ``Are word
  embedding-based features useful for sarcasm detection?'' in \emph{Proceedings
  of the 2016 Conference on Empirical Methods in Natural Language Processing},
  2016, pp. 1006--1011.

\bibitem{joshi2017automatic}
A.~Joshi, P.~Bhattacharyya, and M.~J. Carman, ``Automatic sarcasm detection: A
  survey,'' \emph{ACM Computing Surveys (CSUR)}, vol.~50, no.~5, pp. 1--22,
  2017.

\bibitem{kim2014convolutional}
Y.~Kim, ``Convolutional neural networks for sentence classification,'' in
  \emph{Proceedings of the 2014 Conference on Empirical Methods in Natural
  Language Processing (EMNLP)}, 2014, pp. 1746--1751.

\bibitem{hochreiter1997long}
S.~Hochreiter and J.~Schmidhuber, ``Long short-term memory,'' \emph{Neural
  Computation}, vol.~9, no.~8, pp. 1735--1780, 1997.

\bibitem{scarselli2008graph}
F.~Scarselli, M.~Gori, A.~C. Tsoi, M.~Hagenbuchner, and G.~Monfardini, ``The
  graph neural network model,'' \emph{IEEE Transactions on Neural Networks},
  vol.~20, no.~1, pp. 61--80, 2008.

\bibitem{defferrard2016convolutional}
M.~Defferrard, X.~Bresson, and P.~Vandergheynst, ``Convolutional neural
  networks on graphs with fast localized spectral filtering,'' in
  \emph{Advances in Neural Information Processing Systems}, 2016, pp.
  3844--3852.

\bibitem{bastings2017graph}
J.~Bastings, I.~Titov, W.~Aziz, D.~Marcheggiani, and K.~Sima’an, ``Graph
  convolutional encoders for syntax-aware neural machine translation,'' in
  \emph{Proceedings of the 2017 Conference on Empirical Methods in Natural
  Language Processing}, 2017, pp. 1957--1967.

\bibitem{battaglia2016interaction}
P.~Battaglia, R.~Pascanu, M.~Lai, D.~J. Rezende \emph{et~al.}, ``Interaction
  networks for learning about objects, relations and physics,'' in
  \emph{Advances in Neural Information Processing Systems}, 2016, pp.
  4502--4510.

\bibitem{yao2019graph}
L.~Yao, C.~Mao, and Y.~Luo, ``Graph convolutional networks for text
  classification,'' in \emph{Proceedings of the AAAI Conference on Artificial
  Intelligence}, vol.~33, 2019, pp. 7370--7377.

\bibitem{mukku2017actsa}
S.~S. Mukku and R.~Mamidi, ``Actsa: Annotated corpus for telugu sentiment
  analysis,'' in \emph{Proceedings of the First Workshop on Building
  Linguistically Generalizable NLP Systems}, 2017, pp. 54--58.

\bibitem{mukku2017tag}
S.~S. Mukku, S.~R. Oota, and R.~Mamidi, ``Tag me a label with multi-arm: Active
  learning for telugu sentiment analysis,'' in \emph{International Conference
  on Big Data Analytics and Knowledge Discovery}.\hskip 1em plus 0.5em minus
  0.4em\relax Springer, 2017, pp. 355--367.

\bibitem{schlichtkrull2018modeling}
M.~Schlichtkrull, T.~N. Kipf, P.~Bloem, R.~Van Den~Berg, I.~Titov, and
  M.~Welling, ``Modeling relational data with graph convolutional networks,''
  in \emph{European Semantic Web Conference}.\hskip 1em plus 0.5em minus
  0.4em\relax Springer, 2018, pp. 593--607.

\bibitem{liu2016recurrent}
P.~Liu, X.~Qiu, and X.~Huang, ``Recurrent neural network for text
  classification with multi-task learning,'' in \emph{Proceedings of the
  Twenty-Fifth International Joint Conference on Artificial Intelligence},
  2016, pp. 2873--2879.

\bibitem{battaglia2018relational}
P.~W. Battaglia, J.~B. Hamrick, V.~Bapst, A.~Sanchez-Gonzalez, V.~Zambaldi,
  M.~Malinowski, A.~Tacchetti, D.~Raposo, A.~Santoro, R.~Faulkner
  \emph{et~al.}, ``Relational inductive biases, deep learning, and graph
  networks,'' \emph{arXiv preprint arXiv:1806.01261}, 2018.

\bibitem{tayal2019short}
K.~Tayal, N.~Rao, S.~Agrawal, and K.~Subbian, ``Short text classification using
  graph convolutional network,'' in \emph{NIPS workshop on Graph Representation
  Learning}, 2019.

\bibitem{marcheggiani2017encoding}
D.~Marcheggiani and I.~Titov, ``Encoding sentences with graph convolutional
  networks for semantic role labeling,'' in \emph{Proceedings of the 2017
  Conference on Empirical Methods in Natural Language Processing}, 2017, pp.
  1506--1515.

\bibitem{zhang2020every}
Y.~Zhang, X.~Yu, Z.~Cui, S.~Wu, Z.~Wen, and L.~Wang, ``Every document owns its
  structure: Inductive text classification via graph neural networks,''
  \emph{arXiv preprint arXiv:2004.13826}, 2020.

\bibitem{kipf2016semi}
T.~N. Kipf and M.~Welling, ``Semi-supervised classification with graph
  convolutional networks,'' \emph{arXiv preprint arXiv:1609.02907}, 2016.

\bibitem{kingma2013auto}
D.~P. Kingma and M.~Welling, ``Auto-encoding variational bayes,'' \emph{arXiv
  preprint arXiv:1312.6114}, 2013.

\bibitem{kipf2016variational}
T.~N. Kipf and M.~Welling, ``Variational graph auto-encoders,'' \emph{arXiv
  preprint arXiv:1611.07308}, 2016.

\bibitem{zhang2015character}
X.~Zhang, J.~Zhao, and Y.~LeCun, ``Character-level convolutional networks for
  text classification,'' in \emph{Advances in Neural Information Processing
  Systems}, 2015, pp. 649--657.

\bibitem{caruana1997multitask}
R.~Caruana, ``Multitask learning,'' \emph{Machine Learning}, vol.~28, no.~1,
  pp. 41--75, 1997.

\bibitem{xiao2018gated}
L.~Xiao, H.~Zhang, and W.~Chen, ``Gated multi-task network for text
  classification,'' in \emph{Proceedings of the 2018 Conference of the North
  American Chapter of the Association for Computational Linguistics: Human
  Language Technologies, Volume 2 (Short Papers)}, 2018, pp. 726--731.

\bibitem{parupalli2018enrichment}
S.~Parupalli and N.~Singh, ``Enrichment of ontosensenet: Adding a
  sense-annotated telugu lexicon,'' \emph{arXiv preprint arXiv:1804.02186},
  2018.

\bibitem{gangula2018resource}
R.~R.~R. Gangula and R.~Mamidi, ``Resource creation towards automated sentiment
  analysis in telugu (a low resource language) and integrating multiple domain
  sources to enhance sentiment prediction,'' in \emph{Proceedings of the
  Eleventh International Conference on Language Resources and Evaluation (LREC
  2018)}, 2018.

\bibitem{varshit2018sentiment}
B.~Varshit, B.~V. Vishal, D.~M. M.~K. Reddy, and R.~Mamidi, ``Sentiment as a
  prior for movie rating prediction,'' in \emph{Proceedings of the 2nd
  International Conference on Innovation in Artificial Intelligence}, 2018, pp.
  148--153.

\bibitem{tummalapalli2018towards}
M.~Tummalapalli, M.~Chinnakotla, and R.~Mamidi, ``Towards better sentence
  classification for morphologically rich languages,'' 2018.

\bibitem{choudhary2018sentiment}
N.~Choudhary, R.~Singh, I.~Bindlish, and M.~Shrivastava, ``Sentiment analysis
  of code-mixed languages leveraging resource rich languages,'' \emph{arXiv
  preprint arXiv:1804.00806}, 2018.

\bibitem{damani2013improving}
O.~P. Damani, ``Improving pointwise mutual information (pmi) by incorporating
  significant co-occurrence,'' in \emph{Proceedings of the Seventeenth
  Conference on Computational Natural Language Learning}, 2013, pp. 20--28.

\bibitem{wallach2006topic}
H.~M. Wallach, ``Topic modeling: beyond bag-of-words,'' in \emph{Proceedings of
  the 23rd international conference on Machine learning}, 2006, pp. 977--984.

\bibitem{ramos2003using}
J.~Ramos \emph{et~al.}, ``Using tf-idf to determine word relevance in document
  queries,'' in \emph{Proceedings of the first instructional conference on
  machine learning}, vol. 242.\hskip 1em plus 0.5em minus 0.4em\relax
  Piscataway, NJ, 2003, pp. 133--142.

\bibitem{tang2016effective}
D.~Tang, B.~Qin, X.~Feng, and T.~Liu, ``Effective lstms for target-dependent
  sentiment classification,'' in \emph{Proceedings of COLING 2016, the 26th
  International Conference on Computational Linguistics: Technical Papers},
  2016, pp. 3298--3307.

\bibitem{perozzi2014deepwalk}
B.~Perozzi, R.~Al-Rfou, and S.~Skiena, ``Deepwalk: Online learning of social
  representations,'' in \emph{Proceedings of the 20th ACM SIGKDD International
  Conference on Knowledge Discovery and Data Mining}, 2014, pp. 701--710.

\bibitem{grover2016node2vec}
A.~Grover and J.~Leskovec, ``node2vec: Scalable feature learning for
  networks,'' in \emph{Proceedings of the 22nd ACM SIGKDD International
  Conference on Knowledge Discovery and Data Mining}, 2016, pp. 855--864.

\bibitem{zhao2020summpip}
J.~Zhao, M.~Liu, L.~Gao, Y.~Jin, L.~Du, H.~Zhao, H.~Zhang, and G.~Haffari,
  ``Summpip: Unsupervised multi-document summarization with sentence graph
  compression,'' in \emph{Proceedings of the 43rd International ACM SIGIR
  Conference on Research and Development in Information Retrieval}, 2020, pp.
  1949--1952.

\bibitem{kingma2014adam}
D.~P. Kingma and J.~Ba, ``Adam: A method for stochastic optimization,''
  \emph{arXiv preprint arXiv:1412.6980}, 2014.

\bibitem{mikolov2013distributed}
T.~Mikolov, I.~Sutskever, K.~Chen, G.~S. Corrado, and J.~Dean, ``Distributed
  representations of words and phrases and their compositionality,'' in
  \emph{Advances in Neural Information Processing Systems}, 2013, pp.
  3111--3119.

\end{thebibliography}
\bibliographystyle{IEEEtran}

\end{document}